\documentclass[letterpaper]{article} 
\usepackage{aaai24}  
\usepackage{times}  
\usepackage{helvet}  
\usepackage{courier}  
\usepackage[hyphens]{url}  
\usepackage{graphicx} 
\usepackage{caption} 
\usepackage{algorithm}
\usepackage{algorithmic}

\usepackage{newfloat}
\usepackage{listings}
\DeclareCaptionStyle{ruled}{labelfont=normalfont,labelsep=colon,strut=off} 
\floatstyle{ruled}
\newfloat{listing}{tb}{lst}{}
\floatname{listing}{Listing}
\usepackage[utf8]{inputenc} 
\usepackage[T1]{fontenc}    
\usepackage[pagebackref=false,breaklinks=true,letterpaper=true,colorlinks,bookmarks=false,linkcolor=blue]{hyperref}
\usepackage{url}            
\usepackage{booktabs}       
\usepackage{amsfonts}       
\usepackage{nicefrac}       
\usepackage{microtype}      
\usepackage{lipsum}
\usepackage{graphicx}
\graphicspath{ {./images/} }
\usepackage{caption}

\usepackage{amsmath,amsfonts}
\usepackage{algorithmic}
\usepackage{algorithm}
\usepackage{array}
\usepackage{textcomp}
\usepackage{stfloats}
\usepackage{url}
\usepackage{multirow}
\usepackage{verbatim}
\usepackage{amssymb}
\usepackage{graphicx}
\usepackage{cite}
\usepackage{xcolor}
\usepackage{threeparttable}
\title{Link-Context Learning for Multimodal LLMs}
\author {
    Yan Tai\textsuperscript{\rm 1, \rm 2}$^*$,
    Weichen Fan\textsuperscript{\rm 1}$^*$$^\dag$,
    Zhao Zhang\textsuperscript{\rm 1},
    Feng Zhu\textsuperscript{\rm 1},
    Rui Zhao\textsuperscript{\rm 1},
    Ziwei Liu\textsuperscript{\rm 3}
}
\affiliations {
    \textsuperscript{\rm 1}SenseTime Research\\
    \textsuperscript{\rm 2}Institute of Automation, CAS\\
    \textsuperscript{\rm 3}S-Lab, Nanyang Technological University\\
}

\begin{document}

\twocolumn[{%
\renewcommand\twocolumn[1][]{#1}%
\maketitle

\begin{center}
    \captionsetup{type=figure}
    \includegraphics[width=0.9\linewidth]{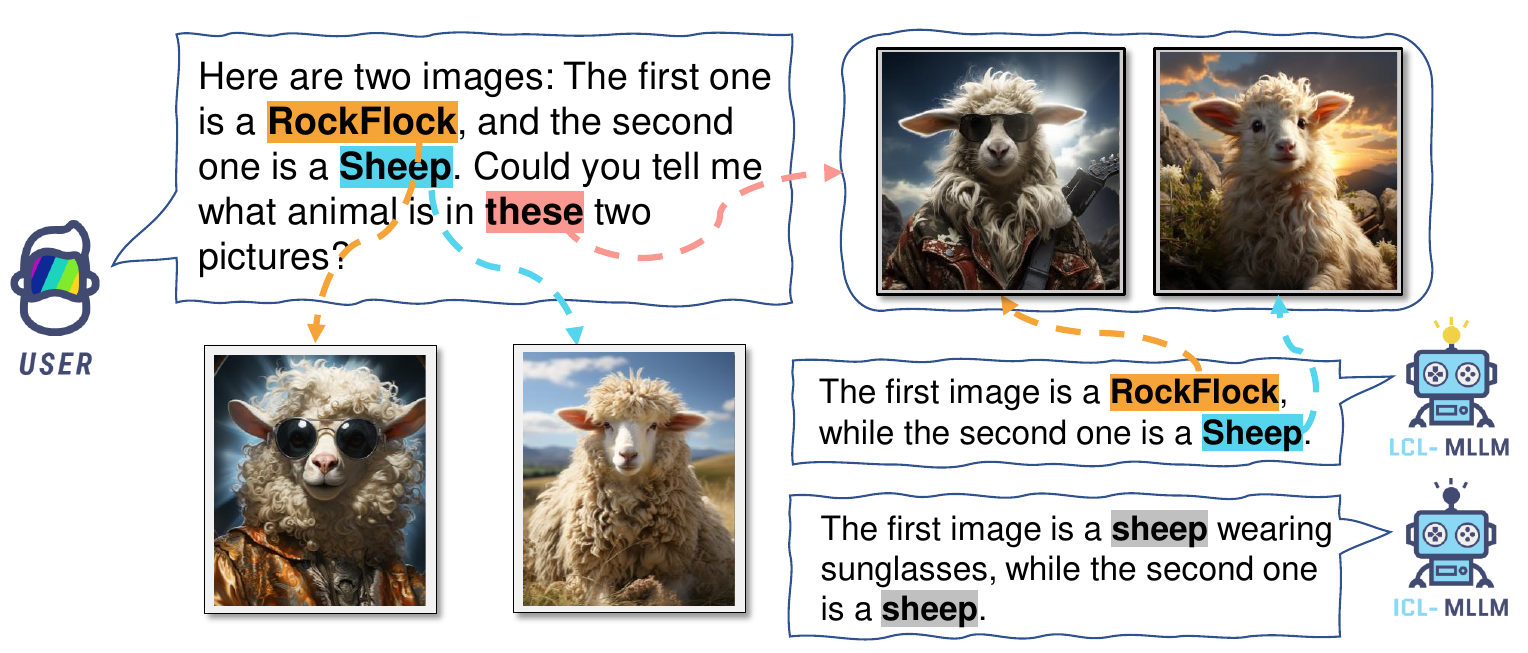}
    \captionof{figure}{\textbf{The demo dialogue of our proposed link-context learning}. After presenting the model with a pair of unseen images and novel concepts, our improved model gains the ability to learn and retain the acquired knowledge throughout the conversation while the vanilla MLLMs fail to provide accurate answers.}
    \label{fig:demo_dia}
\end{center}}]
\renewcommand{\thefootnote}{\fnsymbol{footnote}} 
\footnotetext[1]{Equal Technical Contribution.} 
\footnotetext[2]{Project Lead.} 
\begin{abstract}
The ability to learn from context with novel concepts, and deliver appropriate responses are essential in human conversations. Despite current Multimodal Large Language Models (MLLMs) and Large Language Models (LLMs) being trained on mega-scale datasets, recognizing unseen images or understanding novel concepts in a training-free manner remains a challenge.  In-Context Learning (ICL) explores training-free few-shot learning, where models are encouraged to ``learn to learn" from limited tasks and generalize to unseen tasks. In this work, we propose \textbf{link-context learning (LCL)}, which emphasizes "reasoning from cause and effect" to augment the learning capabilities of MLLMs. LCL goes beyond traditional ICL by explicitly strengthening the causal relationship between the support set and the query set. By providing demonstrations with causal links, LCL guides the model to discern not only the analogy but also the underlying causal associations between data points, which empowers MLLMs to recognize unseen images and understand novel concepts more effectively. To facilitate the evaluation of this novel approach, we introduce the \textbf{ISEKAI} dataset, comprising exclusively of unseen generated image-label pairs designed for link-context learning. Extensive experiments show that our LCL-MLLM exhibits strong link-context learning capabilities to novel concepts over vanilla MLLMs. Code and data will be released at \url{https://github.com/isekai-portal/Link-Context-Learning}.
\end{abstract}

\section{Introduction}
\textit{(In the near future, mankind finally be able to travel interstellar and come to the centaur constellation.)}\\
\textit{\textbf{Human} and \textbf{MLLM} walk off the spaceship.} \\
\textit{\textbf{Human}:``We made it! Look! The locals are here.''}\\
\textit{\textbf{Locals}: Greetings, you can call us 'RockFlock'.}\\
\textit{\textbf{MLLM}: ``Hi, sheep!''} \\
\textit{\textbf{Human}: ``\includegraphics[height=12pt]{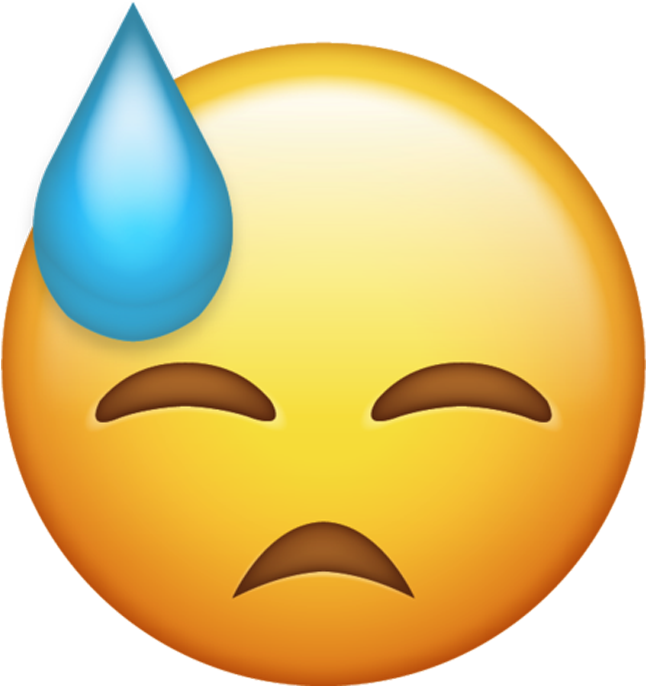}''} \\

The above conversation between humans and MLLMs serves as a humorous representation of how MLLMs struggle to learn from demonstration during the conversation for real. 'RockFlock' is our hand-made species, which possesses both a human-like body and a sheep-like head, as shown in Figure \ref{fig:demo_dia}. Current MLLMs fail to link the unseen image-label pairs to recognize novel objects in a single conversation. To address this limitation, equipping the model with few-shot learning ability has been a long-standing topic in computer vision even before the era of MLLMs. This approach enables the model to learn from limited examples and mitigate the issue effectively. The primary method for MLLMs to learn from demonstrations is known as in-context learning, wherein the models show remarkable improvement on downstream tasks after being exposed to a few input-label pairs.
\begin{figure}[t]
  \centering
  \includegraphics[width=1.\linewidth]{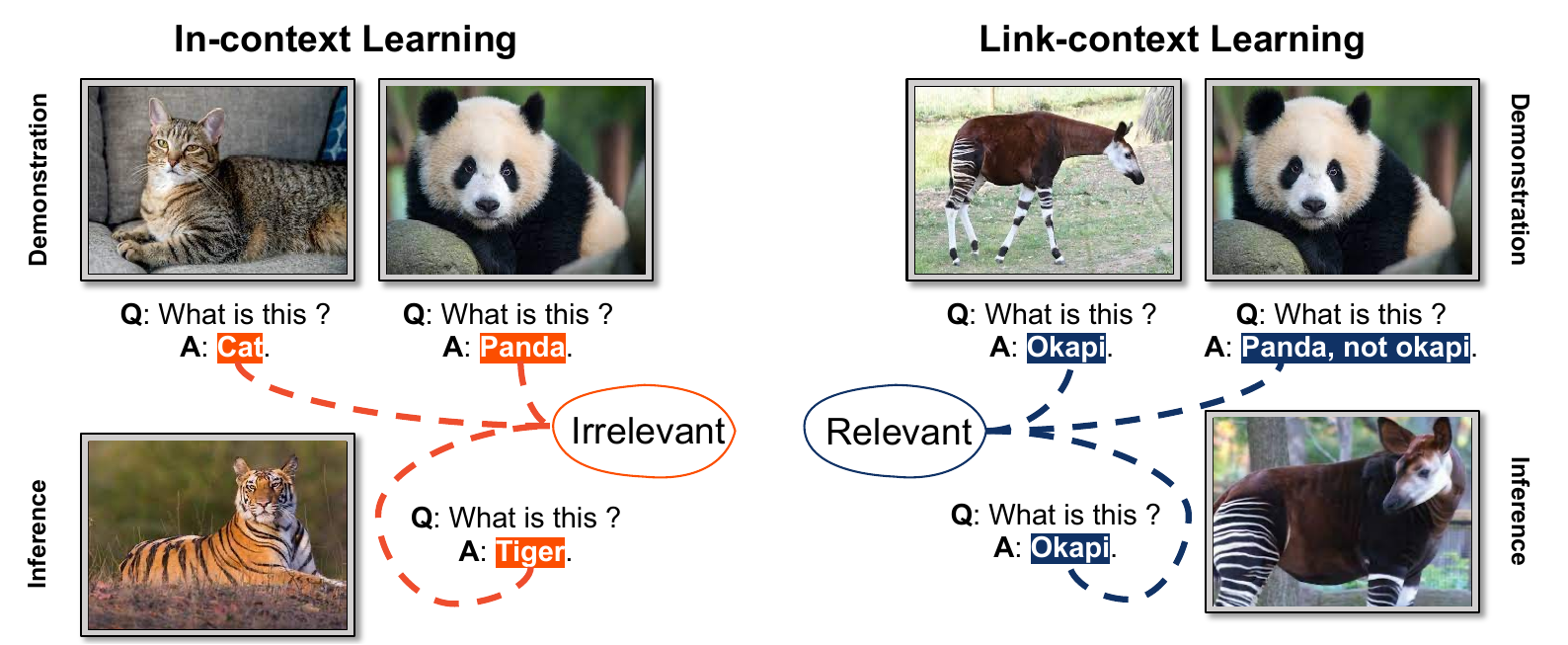}
  \caption{\textbf{The difference between our link-context learning with in-context learning}. In-context learning involves providing irrelevant tasks for demonstration, whereas there is a direct causal relationship between the demonstration and inference phases of link-context learning.
    }
  \label{fig:icl-with-lcl}
\end{figure}
However, current MLLMs have very limited benefits from in-context learning, since the emphasis is primarily on guiding the model to acquire the ability to process novel tasks after ``learning'' from meta tasks. However, the model's performance is not affected even if the answers provided in the meta-tasks are all wrong.~\cite{min2022rethinking} Thus, what MLLMs have ``learned'' from demonstration remains on answering questions in a specific format rather than understanding the causal relationship between the image-label pairs. To enable MLLMs to concentrate more on the causal relationship between the image and label pairs, \textit{Frozen} method~\cite{tsimpoukelli2021multimodal} binds different labels to known images. However, a significant challenge arises when MLLMs encounter entirely novel scenarios where both the image and the label are unseen. In such instances, the task of extracting the underlying cause and effect from the demonstration and making accurate predictions based on this newfound knowledge remains an unsolved puzzle. The 'RockFlock' (unseen images and novel concepts), shown in Figure \ref{fig:demo_dia}, would be misrecognized by the previous methods, while our model learns the concept of 'RockFlock' from the demonstration and makes responses accurately. Moreover, the acquisition of novel concepts does not impede the existing knowledge, enabling the model to effectively distinguish between the original and newly learned images.

Inspired by in-context learning (hereinafter called \textbf{ICL}), we propose \textit{link-context learning} (hereinafter called \textbf{LCL}), which requires the MLLMs to acquire knowledge about new concepts from the conversation and retain their existing knowledge for accurate question-answering. As shown in Figure \ref{fig:icl-with-lcl}, current in-context learning in MLLMs emphasizes benefiting from the causal-irrelevant demonstration. However, for link-context learning, the demonstration and the final task are linked causally.  (e.g. If the 'apple' is renamed as 'orange' in the demonstration, the model should call apple an 'orange' during the inference.) With this ability, the MLLMs could support few-shot learning in a flexible way.

\begin{figure}[t]
  \centering
  \includegraphics[width=1.0\linewidth]{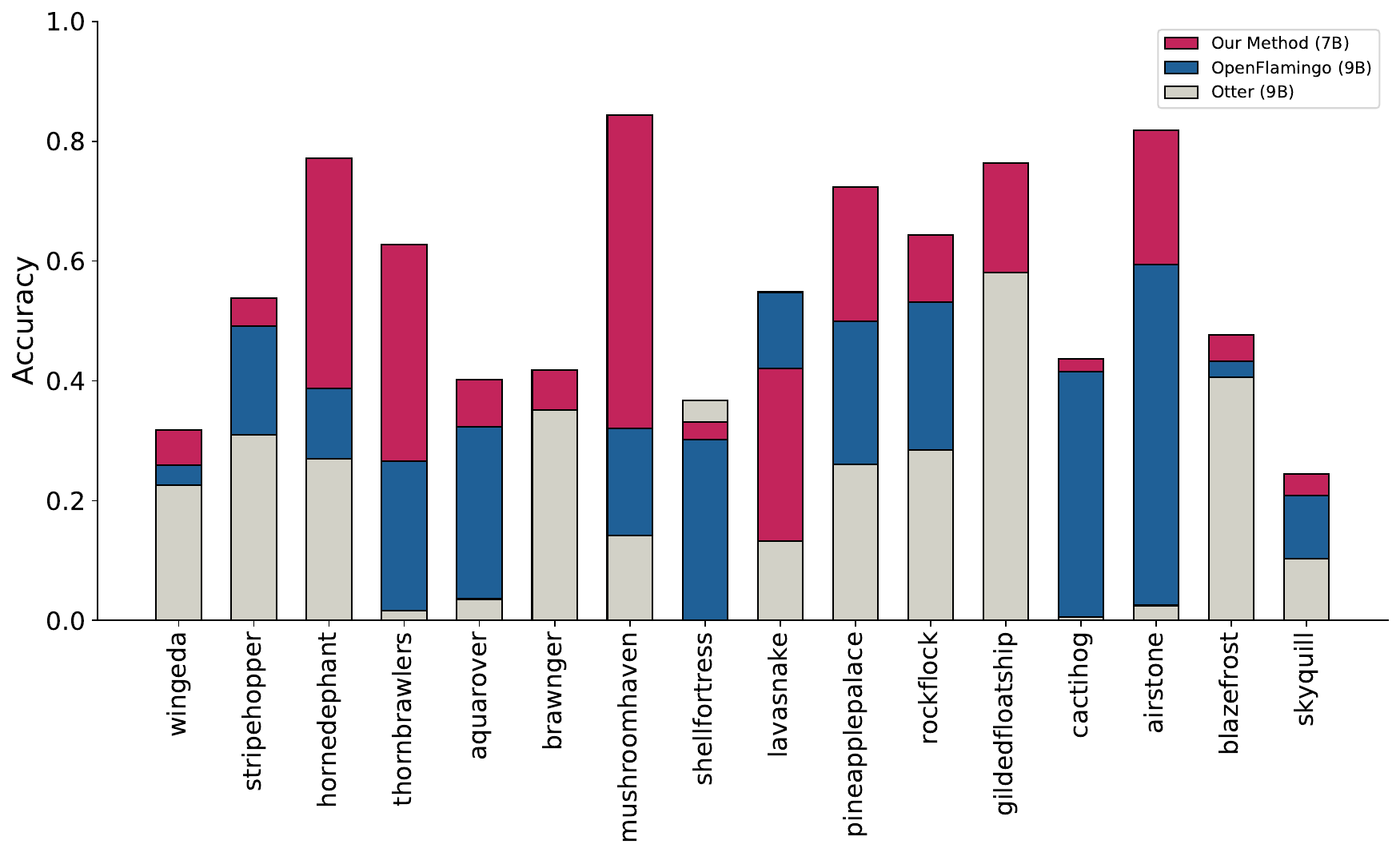}
  \caption{\textbf{Overview of results on several categories of ISEKAI dataset}: Our model outperforms OpenFlamingo (9B)~\cite{anas_awadalla_2023_7733589} and Otter (9B)~\cite{li2023otter} across almost all the categories, showcasing superior performance in scenarios involving entirely unseen images.
    }
  \label{fig:lcl_improve}
\end{figure}

In the era of Large Language Models, evaluating models' performance on few-shot learning becomes a challenge, as these models are extensively trained on vast amounts of real-life data. To address this issue and provide a comprehensive assessment of \textit{link-context learning}, we introduce the \textit{ISEKAI} dataset. This dataset comprises unseen images and concepts, entirely novel to MLLMs, as they transcend the boundaries of realism. All the images in the dataset are generated by Stable Diffusion~\cite{rombach2022high} and Midjourney~\cite{midjourney}, while all the labels or concepts are fabricated as well. Figure~\ref{fig:lcl_improve} shows the comparisons between our model and Otter~\cite{li2023otter}, OpenFlamingo~\cite{anas_awadalla_2023_7733589} on ISEKAI dataset.

In this paper, we present \textit{link-context learning} (LCL), a setting that bestows MLLMs with the capability to understand the potential causal relationship in the conversation and process unseen images and concepts. Unlike ICL mainly focuses on inspiring models with a wide variety of different tasks, LCL goes a step further by empowering the model to establish a mapping between the source and target, thereby enhancing its overall performance. The contributions of this work can be summarized as follows:
\begin{itemize}
    \item \textbf{Link-Context Learning}:
    We introduce a novel causal-relevant few-shot learning setting, where MLLMs are challenged to assimilate new concepts from the ongoing conversation and retain this knowledge for accurate question-answering. Under link-context learning, we empower the MLLMs to grasp the causal relationship between the source and target from the demonstration.
    \item \textbf{ISEKAI Dataset}: 
    Since most real-world data is not completely unseen to MLLMs, we release a challenging fabricated dataset to the public, where novel image-concept pairs are introduced, for evaluation of MLLMs' performance. 
\end{itemize}

\section{Related Works}

Multimodal Large Language Models~\cite{openai2023gpt4,li2022blip,li2023blip2,peng2023vicuna,liu2023llava} have demonstrated significant capabilities in universal generation or recognition tasks. Following the new paradigm of MLLMs, various visual tasks can be achieved in a training-free zero-shot manner~\cite{radford2021clip,li2022glip}, escaping from the heavy \textit{pretrain-and-finetune} process. However, recognize arbitrary content through a single model is generally considered extremely difficult. How to enhancing recognition capability of MLLMs in the wild at a low cost has emerged as a recent research focus.


\paragraph{Multimodal Prompt Tuning} Multimodal Prompt Tuning (M-PT) is commonly used in contrastive learning-based multimodal large models, such as CLIP~\cite{radford2021clip}. In the training process, prompt tuning usually freezes most of the model's parameters and only updates a small number of parameters to achieve results similar to fine-tuning~\cite{yang2022prompt, zhou2022CoOp, zhou2022CoCoOp, liu2023xuejing}. PT~\cite{yang2022prompt} add tunable prompt embeddings to each layer of the encoder and decoder, only the weights of the added embeddings will be updated during training. VPT~\cite{jia2022vpt} added a set of learnable parameters in specific positions to tune the model. CoOp~\cite{zhou2022CoOp} and UPT~\cite{zang2022upt} used CLIP as the backbone and prompted it to fit few-shot settings. CoCoOp~\cite{zhou2022CoCoOp}, POMP~\cite{ren2023pomp} and MaPLe~\cite{khattak2023maple} extend prompt tuning to open-vocabulary visual recognition tasks. However, traditional prompt tuning methods are not suitable for the powerful generative multimodal large language models.

\paragraph{Multimodal Instruction Tuning} Multimodal Instruction Tuning (M-IT) enhances the zero-shot capability of MLLMs in unseen tasks by fine-tuning them on an instruction descriptions-based dataset ~\cite{wei2022flan, li2022blip, liu2023llava, openai2023chatgpt, openai2023gpt4}. MiniGPT-4~\cite{zhu2023minigpt4} and LLaVA~\cite{liu2023llava} keep the visual encoder frozen and tune the language model, extending instruction tuning to multimodality. mPLUG-Owl~\cite{ye2023mplugowl} tuned visual and text encoder separately in two stages, and proposed an evaluation dataset for assessing vision-related instruction tuning. InstructBLIP~\cite{dai2023instructblip} enhances zero-shot capability by performing instruction tuning on multiple datasets. Shikra~\cite{chen2023shikra} and Kosmos-2~\cite{peng2023kosmos2} expanded MLLMs to visual grounding tasks using instructions with bounding box coordinates. Even though these studies demonstrate outstanding zero-shot capability, they still cannot recognize classes that were not seen during the model training process.

\paragraph{Multimodal In-Context Learning} Large Language Models (LLMs) have shown outstanding capability in learning from context samples. In the Multimodal In-Context Learning (M-ICL) settings, following the input image samples and optional instruction, MLLMs can learn new task patterns in a few-shot manner~\cite{dong2023survey, yang2023mmreact, lu2023chameleon, gupta2022visual}. Flamingo~\cite{NEURIPS2022flamingo} takes in-context learning into consideration during the pretraining process, allowing the model to possess the ability to support in-context learning. Otter~\cite{li2023otter} follows Flamingo and proposed a new in-context learning dataset, proceeding with the ICL capability in the instruction tuning stage. 

Different from previous methods, our proposed \textit{link-context learning} can establish a causal link between the support and query set. Specifically, using few-shot class-specific images and textual prompts, LCL can link the prompt and inference samples, and even associate previously unseen images with new concepts.


\section{Link-Context Learning}
In this section, we first give a brief introduction to in-context learning and unveil its main restrictions and difference to our link-context learning in \nameref{sec:preliminary}; next, we bring the power of link-context learning into MLLMs in \nameref{sec:method}.

\subsection{Preliminary}\label{sec:preliminary}
\noindent\textbf{In-Context Learning}
Formally, in-context learning~\cite{brown2020language} refers to: the model should choose the answer with the highest prediction score from a set candidate answers $Y = \{y_1,y_2,...,y_n\}$, given a query input $x$, conditioning on a support set $S$, which consists of multiple input-label pairs from a wide variety of tasks, where $S = \{(x_1,y_1),(x_2,y_2),...,(x_n,y_n)\}$. (The query and the sample of $S$ should belong to different tasks.) 

From another perspective, in-context learning could be denoted as training-free few-shot learning, as it transforms the training stage of few-shot learning into the demonstration input for Large Language Models. Noted that the ICL~\cite{brown2020language} is consistent with FSL, where the tasks in the demonstration (training) stage and in the inference (query) stage are different.




\noindent\textbf{Link-Context Learning}
Essentially, link-context learning (LCL) represents a form of training-free and causal-linked few-shot learning. In this approach, a support set $S = {(x_1,y_1),(x_2,y_2),...,(x_n,y_n)}$ is provided, along with a query sample $x$ from the query set $Q$, where the data pairs from the support set are causally linked to the query set. The model is tasked with predicting the answer based on the causal-linked relationship between the query and support set.

To provide further clarity, link-context learning significantly strengthens the causal relationship between the support set and the query set. For instance: 1). New arithmetic rules: In this scenario, the support set consists of arithmetic expressions such as ${(1\ \text{<op>} \ 2=3),(2\ \text{<op>}\ 3=5)}$, with the query sample being $4\ \text{<op>}\ 5=?$. Here, "<op>" represents a new arithmetic rule that we aim to teach the model through the demonstration; 2). Novel image classification: In this case, the support set contains pairs like ${(\text{<unseen image>}:\text{<novel cls A>})}$, ${(\text{<unseen image>}:\text{<novel cls B>})}$, while the query sample is $(\text{<unseen image>}\ \text{belongs to?})$. This example demonstrates how we expect the model to correctly classify the unseen image into one of the specified novel classes based on the demonstration.

In essence, link-context learning enhances the model's capacity to grasp new concepts and relationships by effectively establishing a causal link between the support set and the query set. While this setting is applicable to both LLMs and MLLMs, our primary focus in this paper is on the application of link-context learning specifically in MLLMs. By concentrating on MLLMs, we aim to showcase the potential of this approach in multimodal models and its implications for advancing their learning capabilities.


\subsection{Bring Link-Context Learning to MLLMs} \label{sec:method}
In this section, our main objective is to introduce Link-Context Learning (LCL) to the realm of MLLMs. Recognizing that the current MLLMs trained in the ICL manner may not excel in LCL tasks, we propose a novel training strategy to fine-tune MLLMs. This approach aims to equip the models with the capability to grasp causal links from context effectively. By leveraging this novel training strategy, we aim to empower MLLMs to excel in tasks that require reasoning and understanding causal relationships, thereby broadening their range of capabilities and improving their overall performance. To be more specific, we choose Shikra~\cite{chen2023shikra} as our baseline, and we divide ImageNet1k into ImageNet-900 and ImageNet-100 by classes, which would be discussed in detail in \nameref{sec:LCL-training_dataset}. Additionally, we incorporate the concept of contrast learning in our training strategy, as discussed in \nameref{sec:LCL-training_strategy}. This helps guide the model to understand the shared characteristics among samples of the same kind and the distinctions between samples of different kinds. 

\subsubsection{Training Dataset} \label{sec:LCL-training_dataset}
Unlike traditional tasks that require extensive training data, LCL concentrates on acquiring the ability to  find the link between the source-target pairs in demonstration and generalize to the query samples. Thus, adequate representation of diverse image categories is essential to enable MLLMs to grasp causal relationships effectively and efficiently.

ImageNet1k ~\cite{ILSVRC15IMAGENET} is commonly employed for image classification tasks, and it is customary to train models on the entire dataset to enhance their recognition ability across all categories. In contrast, within the training configuration of LCL, we only select a limited number of samples randomly from each category. Then we arrange a set of related categories with decreasing similarity for each category, referred to as "neighbors". Specifically, we adopted CLIP~ \cite{radford2021clip} to calculate the similarity between different classes within the training dataset. Firstly, we randomly select 100 images from each class and calculate the average image feature for each class. Subsequently, we encode the text names of all classes to obtain their corresponding feature vectors. Ultimately, we compute weighted similarities across distinct class pairs, encompassing image-to-image, image-to-text, and text-to-text correlations. For a specific category, we sort all other categories based on similarity and divide them into $N$ intervals. Then, within each interval, we randomly select categories to construct a set of "neighbors" with a total quantity of $N$.

\subsubsection{Training Strategy} \label{sec:LCL-training_strategy}
In order to make MLLMs understand the causal link between the support set and query sample, as well as the causal relationship between the input-label pairs in the support set, we build positive-negative pairs to urge the model to learn from comparisons. Let the support set be denoted as $S=\{s_1,s_2,...,s_n\}$. Based on the correlation among its samples, we can redefine the support set as $C=\{c_1,c_2,...,c_m\}$, where each $c_m$ serves as a prototype representing a cluster of samples from $S$. These prototypes capture the essential relationships and similarities among samples within $S$. Given the query $x$, we train $\theta$ to maximize the likelihood:
\begin{align}
    \log p_{\theta}(y|x) = \sum_l \log p_{\theta}(y_l|x,C,y_1,y_2,...,y_{l-1}),
\end{align}
where $\theta$ denotes the parameters of the language model. The parameters of the visual encoder are frozen during the training.

\noindent \textbf{[2-way] strategy:}
In this strategy, we train the MLLMs for binary image classification, where the $C=\{c_1,c_2\}$. To be more specific, $c_1$ and $c_2$ here represent the prototype of two classes. We denote the training class set as $T=\{t_1,t_2,...,t_{100}\}$, we randomly sample a class $t_i$ as the positive class, where its neighbor class set $N^{t_i}=\{n^{t_i}_1,n^{t_i}_2,...,n^{t_i}_{100}\}$ ($n^{t_i}_1$ is the most similar class to $t_i$, while the $n^{t_i}_{100}$ is the least). Then we apply a hard-negative mining strategy, where we sample the negative class $n^{t_i}_j$ from $N^{t_i}$ with a probability $p_j=\frac{101-j}{\sum_{m=1}^{100} m}$. Noted that this setting is fixed to train on 16 shots.
\\[1ex]
\noindent \textbf{[2-way-random] strategy:}
In this strategy, we first train the MLLMs on fixed-16 shots following the [\textit{2-way}] strategy, then further train the model with shots averaged sampled from 2-16 shots for 10 epochs.
\\[1ex]
\noindent \textbf{[2-way-weight] strategy:}
Within this strategy, we initially train the MLLMs using a fixed-16 shot regimen, adhering to the [\textit{2-way}] approach. Subsequently, we refine the model by additional training with shots sampled from the range of 2-16, with each shot's probability denoted as $p_j=\frac{e^j}{\sum_{m=2}^{16}e^m}$.
\\[1ex]
\noindent \textbf{[mix] strategy:}
To enhance the model's generalizability, we undertake a fine-tuning process that involves both [2-way] tasks and Shikra's~\cite{chen2023shikra} original tasks. During each iteration, the training samples are evenly sampled from both the [2-way] tasks and the original tasks. This balanced approach ensures that the model gains proficiency in both the newly introduced link-context learning tasks and the pre-existing tasks from Shikra~\cite{chen2023shikra}.


\begin{figure*}[t]
  \centering
  \includegraphics[width=1.0\linewidth]{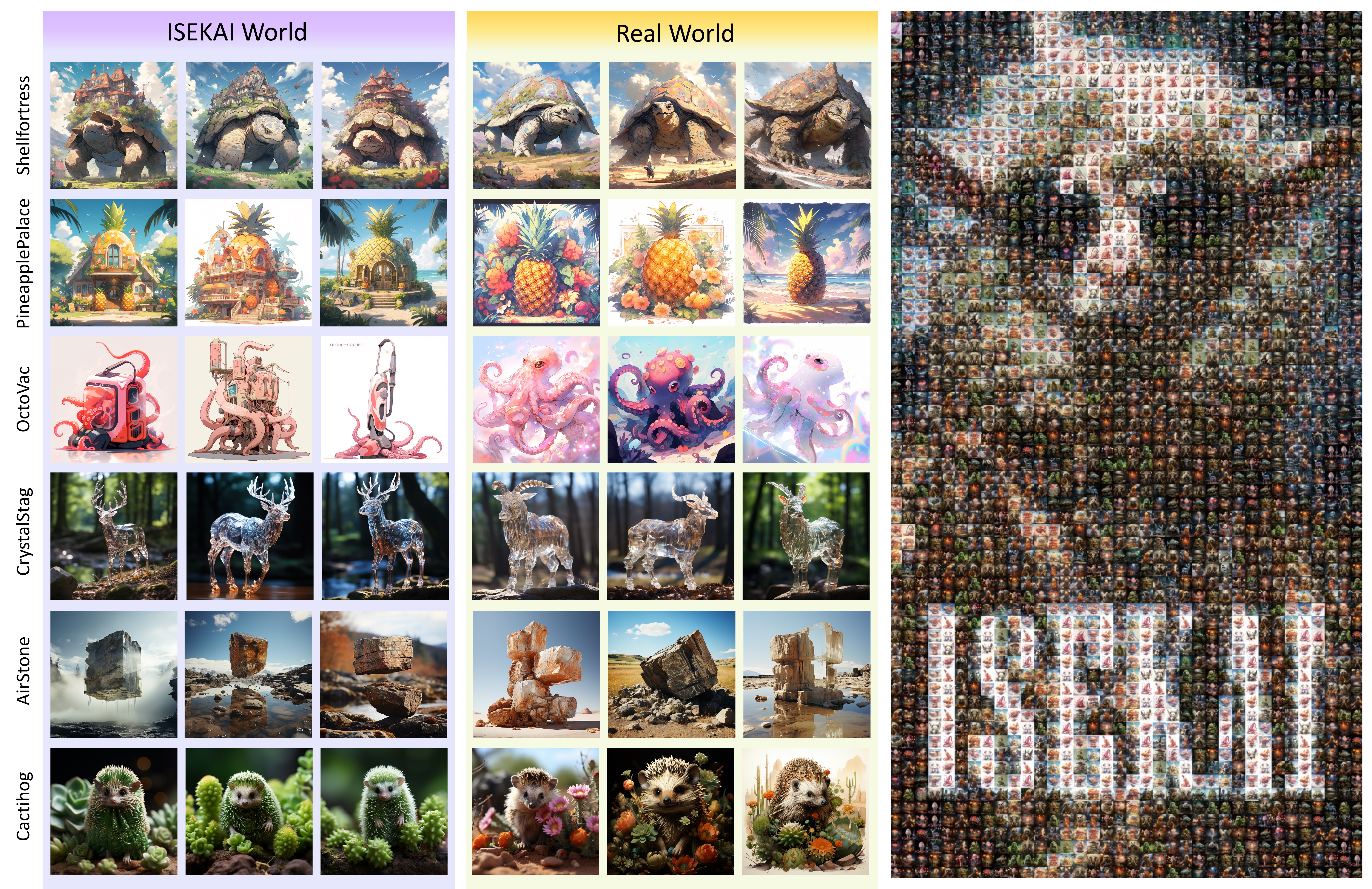}
  \caption{\textbf{Overview of the ISEKAI Dataset}: This dataset comprises entirely generated images, where the images from ``ISEKAI World'' are non-existent in real life, while the images from ``Real World'' are sourced from reality.
    }
  \label{fig:ISEKAI_dataset}
\end{figure*}
\begin{table*}[t]
  \centering
  \resizebox{0.99\textwidth}{!}{
  \begin{tabular}{@{}c|c|c|c|c|c|c|c|c|c}
    \toprule[1.1pt]
     Setting & Method  & 2-shot & 4-shot & 6-shot & 8-shot &10-shot & 12-shot & 14-shot & 16-shot \\
    \hline
    \multirow{5}{*}{ISEKAI-10} & OpenFlamingo~\cite{anas_awadalla_2023_7733589}  & 0.46 & 0.44 & 0.46 & 0.48 & 0.50 & 0.50 & 0.48 & 0.46 \\ 
    \multirow{5}{*} & Otter~\cite{li2023otter}  & 0.23 & 0.23 & 0.19 & 0.15 & 0.14 & 0.12 & 0.10 & 0.07 \\
    \multirow{5}{*} & Vanilla-Shikra~\cite{chen2023shikra} & 0.00 & 0.00 & 0.00 & 0.00 & 0.00 & 0.00 & 0.00 & 0.00\\
    \multirow{5}{*} & Ours-[2-way-random] & \underline{0.64} & \underline{0.63} & \underline{0.65} & \underline{0.62} & \underline{0.61} & \underline{0.57} & \underline{0.56} & \underline{0.56} \\
    \multirow{5}{*} & Ours-[mix] & \textbf{0.68} & \textbf{0.70} & \textbf{0.73} & \textbf{0.69} & \textbf{0.63} & \textbf{0.62} & \textbf{0.65} & \textbf{0.62}\\
    \hline
    \multirow{5}{*}{ISEKAI-pair} & OpenFlamingo~\cite{anas_awadalla_2023_7733589} & 0.19 & 0.34 & 0.38 & 0.39 & \underline{0.41} & \underline{0.40} & \underline{0.40} & \underline{0.40} \\ 
    \multirow{5}{*} & Otter~\cite{li2023otter} & 0.01 & 0.04 & 0.04 & 0.03 & 0.03 & 0.02 & 0.02 & 0.01 \\
    \multirow{5}{*} & Vanilla-Shikra~\cite{chen2023shikra} & 0.00 & 0.00 & 0.00 & 0.00 & 0.00 & 0.00 & 0.00 & 0.00\\
    \multirow{5}{*} & Ours-[mix] & \underline{0.39} & \underline{0.38} & \underline{0.38} & \underline{0.40} & 0.40 & 0.39 & 0.37 & 0.35\\
    \multirow{5}{*} & Ours-[2-way-random] & \textbf{0.43} & \textbf{0.46} & \textbf{0.47} & \textbf{0.48} & \textbf{0.48} & \textbf{0.49} & \textbf{0.49} & \textbf{0.49}\\
    \bottomrule[1.1pt]
  \end{tabular}}
  \caption{\textbf{Quantitative evaluation on ISEKAI} from zero-shot to 16-shot, measured by accuracy. We achieve the best results compared with Otter~\cite{li2023otter} and OpenFlamingo~\cite{anas_awadalla_2023_7733589}.}
  \label{tab:ISEKAI_10}
\end{table*}

\section{ISEKAI Dataset}
To objectively evaluate MLLM's ability to learn new concepts through LCL, we created an ISEKAI dataset, shown in Figure~\ref{fig:ISEKAI_dataset}. The concepts involved are unreal, rarely seen in legends, myths, or fictional media. Thus, MLLM's exposure to these concepts is minimal. The term "Isekai" originates from a fantasy subgenre in anime. Plots usually involve characters transported to a different world, like a fantasy realm or virtual universe. Audiences understand the new world gradually through the protagonist's exploration, akin to MLLM's journey into a new realm of knowledge.

The dataset's images are generated by Midjourney's~\cite{midjourney} text-to-image model using well-crafted instructions. Images were manually selected to ensure core concept consistency. The dataset currently comprises 20 groups, and 40 categories in total (continues to grow). Each group pairs a new concept with a related real-world concept, like "octopus vacuum" and "octopus." These can serve as challenging negative samples for each other. Each concept has no less than 32 images, supporting multi-shot examples. These features enable ISEKAI to comprehensively assess the model's LCL capability. We also provide text descriptions of each concept's appearance and name, contributing to evaluations beyond LCL.

In this paper, we evaluated different models' performance on ISEKAI. For details, refer to \nameref{sec:isekai_res}.

\section{Experiments}
In this section, we present the results of our experiments to showcase the effectiveness of our proposed method. We conduct comprehensive comparisons between our approach (link-context learning-based) and other in-context learning-based MLLMs. 

\begin{figure*}[t]
  \centering
  \includegraphics[width=1.0\linewidth]{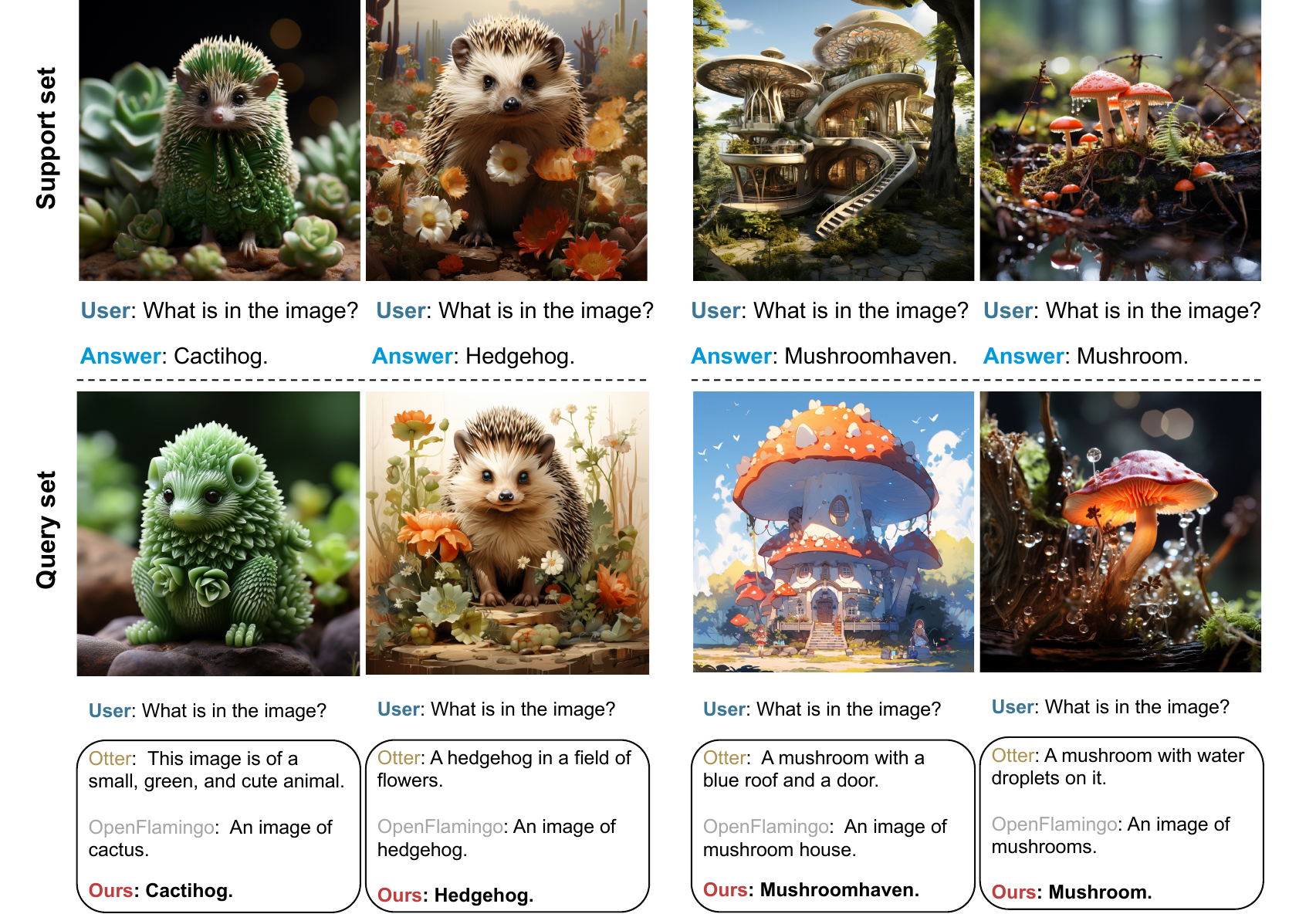}
  \caption{\textbf{Qualitative comparisons} of novel images understanding results between ours and OpenFlamingo~\cite{anas_awadalla_2023_7733589}, Otter~\cite{li2023otter}. The name ``Cactihog'' is a fusion of ``cactus'' and ``hedgehog'', combining the key features of these two creatures. The name ``MushroomHaven'' suggests a dwelling place characterized by giant mushrooms
    }
  \label{fig:ISEKAI_dia}
\end{figure*}
\subsection{Results on ISEKAI}\label{sec:isekai_res}
To quantitatively evaluate the performance of link-context learning, we compare our methods in different strategies with our baseline (Shikra~\cite{chen2023shikra}) as well as ICL methods (Otter and OpenFlamingo) in two challenge datasets: ISEKAI-10 and ISEKAI-pair.\\[1ex]
\noindent \textbf{ISEKAI-10 Evaluation:} Comprising 10 classes of challenging positive-negative image pairs, \textbf{ISEKAI-10} presents a scenario where the positive class is entirely nonexistent in the real world yet shares certain characteristics with the negative class, which comprises common animals or objects from our reality. The upper section of Table~\ref{tab:ISEKAI_10} showcases the outcomes on the ISEKAI-10 dataset, where vanilla-shikra~\cite{chen2023shikra} encountered difficulty. Our model demonstrates competitive performance compared with OpenFlamingo~\cite{anas_awadalla_2023_7733589} and Otter~\cite{li2023otter} across all shot numbers.\\[1ex]
\noindent \textbf{ISEKAI-pair Evaluation:} In the \textbf{ISEKAI-pair} evaluation, positive and negative pairs are constructed using all image categories that do not exist in the real world. Each individual image is paired with all images from other categories, facilitating a comprehensive assessment. This evaluation provides a realistic gauge of the model's capability to handle complete unknowns through various combinations. The lower section of Table~\ref{tab:ISEKAI_10} underscores our model's superiority over OpenFlamingo~\cite{anas_awadalla_2023_7733589} and Otter~\cite{li2023otter} in this context.\\[1ex]
\noindent \textbf{Qualitative Results:} Figure~\ref{fig:demo_dia} provides a visual comparison between our model and OpenFlamingo~\cite{anas_awadalla_2023_7733589}, as well as Otter~\cite{li2023otter}. Notably, our model demonstrates its proficiency in accurately comprehending novel concepts and effectively discerning unfamiliar objects from those with close resemblance. This observation underscores our model's capacity to capture the causal relationship between the source and target domains from the demonstration.

\begin{table*}[t]
  \centering
  \resizebox{0.99\textwidth}{!}{
  \begin{tabular}{@{}c|c|c|c|c|c|c|c|c|c}
    \toprule[1.2pt]
     Method & zero-shot & 2-shot & 4-shot & 6-shot & 8-shot &10-shot & 12-shot & 14-shot & 16-shot \\
    \midrule
    OpenFlamingo~\cite{anas_awadalla_2023_7733589} & 0.00 & 0.41 & 0.62 & 0.72 & 0.75 & 0.77 & 0.78 & 0.73 & 0.72 \\
    Otter~\cite{li2023otter} & 0.13 & 0.18 & 0.21 & 0.24 & 0.25 & 0.26 & 0.24 & 0.23 & 0.23 \\
    \midrule
    Vanilla-Shikra~\cite{chen2023shikra} & \underline{0.05} & 0.0 & 0.0 & 0.0 & 0.0 & 0.0 & 0.0 & 0.0 & 0.0 \\
    Ours-[mix] & \textbf{0.16} & \underline{0.73} & \underline{0.78} & \textbf{0.83} & 0.73 & 0.71 & 0.72 & 0.65 & 0.57 \\
    Ours-[2-way] & 0.02 & 0.51 & 0.61 & 0.68 & 0.73 & 0.77 & \underline{0.78} & \underline{0.78} & \underline{0.79} \\
    Ours-[2-way-random] & 0.0 & \textbf{0.77} & \textbf{0.78} & \underline{0.77} & \textbf{0.79} & \textbf{0.77} & 0.77 & 0.77 & 0.75 \\
    Ours-[2-way-weight] & 0.0 & 0.69 & 0.71 & 0.72 & \underline{0.76} & \underline{0.77} & \textbf{0.78} & \textbf{0.78} &  \textbf{0.79}\\
    \bottomrule[1.2pt]
  \end{tabular}}
  \caption{\textbf{Quantitative evaluation on ImageNet-100} from zero-shot to 16-shot, measured by accuracy. We achieve the best results compared with Otter~\cite{li2023otter} and OpenFlamingo~\cite{anas_awadalla_2023_7733589}.}
  \label{tab:test100}
\end{table*}
\begin{figure*}[t]
\minipage{0.32\textwidth}
  \includegraphics[width=\linewidth]{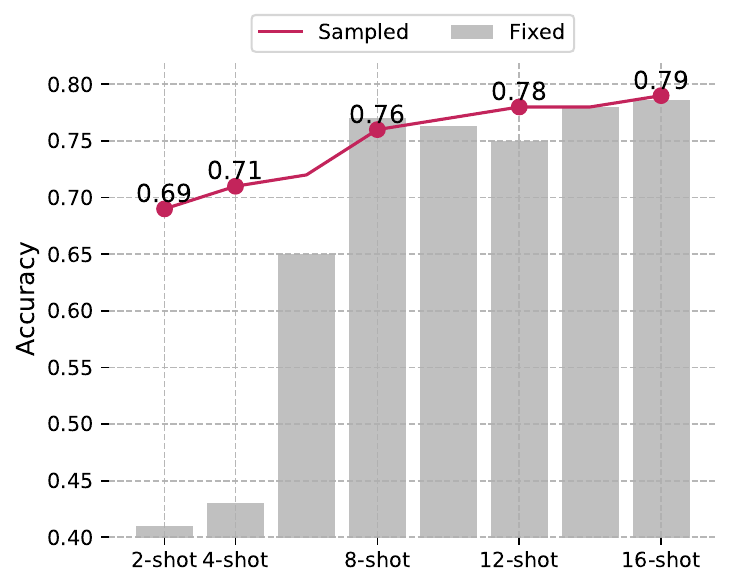}
  \caption{\textbf{The ablation study on shot number}. The grey bars illustrate the highest accuracy achieved for each shot number, denoting specific shot-based training. The red line illustrates the performance of the model trained using a sampled strategy. Notably, both scenarios exhibit plateaus in accuracy after reaching the 8-shot mark.}\label{fig:optimal_shot}
\endminipage\hfill
\minipage{0.32\textwidth}
  \includegraphics[width=\linewidth]{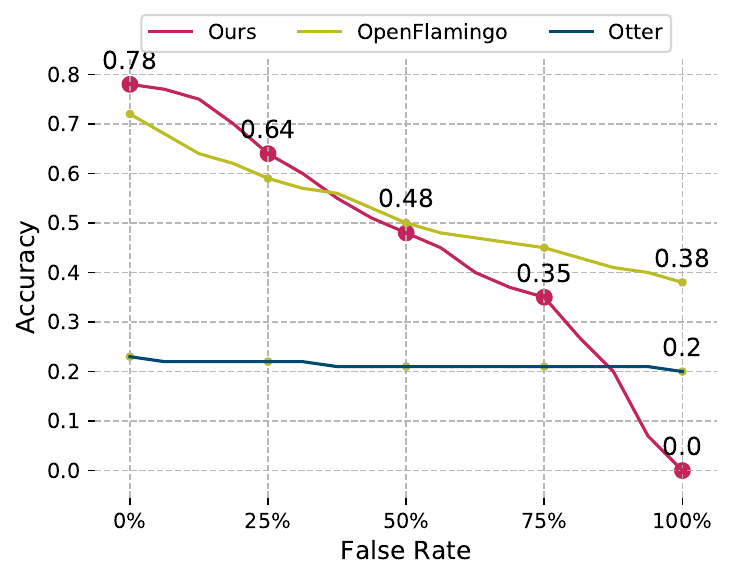}
  \caption{\textbf{The ablation study on false rate}. In contrast to OpenFlamingo~\cite{anas_awadalla_2023_7733589}, which sustains a 38\% accuracy at a 100\% false rate, our model attains 0\% accuracy under the same conditions. This outcome underscores our model's ability to preserve precise linkages between the support set and the query.}\label{fig:false_rate}
\endminipage\hfill
\minipage{0.32\textwidth}%
  \includegraphics[width=\linewidth]{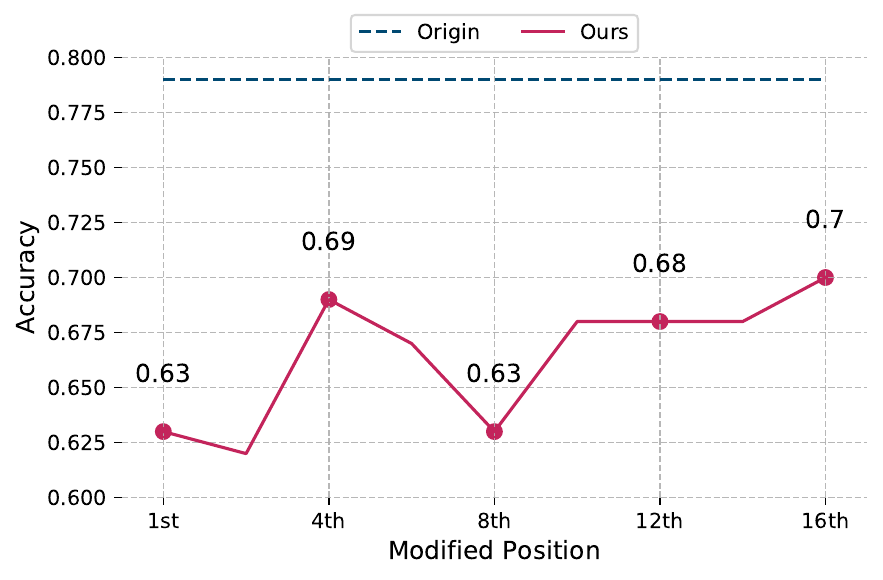}
  \caption{\textbf{The effect of label modifications at distinct positions}. The dashed blue line serves as a reference for the original accuracy, while the red line portrays the accuracy of our model subsequent to the label modified at specific positions. Significant accuracy drop reflects position dependency, while minor change indicates position insignificance in the model's decision-making.}\label{fig:pos}
\endminipage
\end{figure*}
\subsection{Results on ImageNet-100}

We proceed to assess our model's performance on ImageNet-100, encompassing 100 classes that were entirely absent from the training phase. The outcomes underscore the efficacy of our \textit{mix} strategy, which attains the highest accuracy of \textbf{83\%} at 6-shot. In contrast, Otter achieves a peak accuracy of \textbf{25\%}, and OpenFlamingo's performance reaches \textbf{78\%}. Unlike the ISEKAI dataset, the images from ImageNet-100 do correspond to real-world entities.


\subsection{Ablation Study}
\noindent \textbf{Does the ground-truth input-label mapping exists?}

\noindent We conduct an ablation analysis on the correctness of labels within the demonstration (support set). Given a set of image domains ${\mathcal{X}c \in \mathbb{R}^{H\times W\times 3}}$ and label domains ${\mathcal{C}\in \mathbb{R}^N}$, a mapping $f:\mathcal X_c\to\mathcal C$ exists to associate each image with its corresponding label. We use several image-label pairs $\{(x_{c_1}^1,c_1),(x_{c_1}^2,c_1),...,(x_{c_1}^n,c_1)\}$, where $x_{c_i}^{j}\in \mathcal{X}_{c_i}$, as the support set. The model is going to predict the correct answer from a candidate set $Y$:
\begin{align}
    \hat{y} = \mathop{\arg\max}_{y_i\in Y} P(y_i|x,f),
\end{align}
where the prediction is conditioned on the mapping $f$. Consequently, intentionally breaking the mapping relationship within the support set would lead the model to provide incorrect answers, as it heavily relies on the accurate association between the image-label pairs of the support set to make precise predictions. As shown in Figure \ref{fig:false_rate}, we disturb the mapping $f$ by gradually inserting false labels into the support set, and the accuracy falls from $0.78$ to $0.00$ when the correctness of the labels falls from $100\%$ to $0\%$. These results clearly show that maintaining accurate associations between image-label pairs within the support set plays a crucial role in link-context learning.\\[1ex]


\noindent \textbf{Would the model benefit from using a larger shot?}

\noindent Much like supervised learning, the model's accuracy experiences rapid initial growth with an increasing amount of training data, eventually reaching a plateau. During this phase, the selection of more representative samples becomes crucial. Figure \ref{fig:optimal_shot} presents two outcomes: one depicts model accuracy from separate training at a fixed shot (gray bar in the figure), while the other showcases the model's performance through sampling across various shots (red line in the figure). The results reveal slight gains from lower fixed-shot training and consistent performance from random-shot training. Notably, in both random and fixed settings, accuracy plateaus or experiences gradual growth after the 8-shot threshold.\\[1ex]
\noindent \textbf{What does the model's decision-making in the case of multi-shot depend on?}

As shown in Fig~\ref{fig:pos}, when disturbing the label of different positions, the accuracy of the model with 16-shot drops differently, which reflects the extent to which the model prefers different locations. We observe that the model heavily relies on the beginning and the middle positions. From another aspect, it provides an explanation of why the model encounters a plateau in a higher number of shots. Similarly, this phenomenon also exists in LLMs~\cite{liu2023lost}, where the language model tends to be ``lost in the middle'' when processing long contexts. They also reveal that the model's performance keeps decreasing when the contexts grow longer.\\[1ex]
\noindent \textbf{What is the difference between different training strategies?}

\noindent Table~\ref{tab:test100} presents a comprehensive view of the outcomes achieved through our four distinct training strategies. The \textit{mix} strategy stands out by elevating the zero-shot accuracy from 5\% to 16\% and attaining a remarkable 83\% accuracy at 6-shot; however, its performance diminishes to 57\% at 16-shot. In contrast, the \textit{2-way} strategy, anchored at 16-shot training, initiates with a 51\% accuracy at 2-shot and progressively ascends to 79\% at 16-shot. Interestingly, we observe that the accuracy trend of the \textit{2-way} strategy isn't solely attributable to an increase in shots, but rather stems from a closer alignment with the trained pattern. To validate this, we introduce two additional settings: \textit{2-way-random} and \textit{2-way-weight}. These settings undergo fixed-shot training for initialization, followed by finetuning across 2-16 shots with random and weighted approaches, respectively. Both exhibit considerable accuracy improvements in lower shots. Notably, while the accuracy of higher shots, finetuned with a random strategy, drops—an observation mirroring the behavior of the \textit{mix} strategy. These results underscore the efficacy of an even, sustained, and generalized training approach in harnessing the potential of large language models, revealing the emergence of a "lost-in-the-middle" phenomenon, in coherence with our earlier observations.\\[1ex]
\noindent \textbf{Does the training harm the zero-shot performance?}

\noindent Table~\ref{tab:zero-shot-harm} shows the comparison between our-7B model with shikra-13B~\cite{chen2023shikra} and some previous SOTA methods on Imagenet-100 and VQAv2. From the results, we conclude that our \textit{mix} training strategy would not harm the model's zero-shot performance.

\section{Discussion}
\subsection{Limitations}
We believe that this work introduces a challenging and promising setting for both MLLMs and LLMs. However, the primary focus in this paper lies on link-context learning within the context of MLLMs, specifically validating the basic tasks such as image classification. Consequently, this work should be regarded as a foundational baseline for exploring the potential of link-context learning.

Looking ahead, future research directions encompass a deeper theoretical analysis that delves into the intricacies of the causal relationship between the support samples and, crucially, between the support set and the query. Understanding and unraveling the complexities of these causal links represent meaningful avenues of inquiry that could lead to significant advancements in the capabilities of models in reasoning, learning, and adapting to novel scenarios. As the field progresses, we anticipate further investigations and refinements that will not only enrich our understanding of link-context learning but also implement in-context learning for MLLMs and LLMs in a unified way.
\begin{table}[t]
  \centering
  \resizebox{0.47\textwidth}{!}{
  \begin{tabular}{@{}c|c|c|c}
    \toprule[1.2pt]
     Method & ImageNet-100 & $\text{VQAv2}^{\text{dev}}$ & $\text{VQAv2}^{\text{std}}$ \\
    \midrule
    OpenFlamingo~\cite{anas_awadalla_2023_7733589} & 0.00 & - & -\\
    Flamingo-80B~\cite{NEURIPS2022flamingo} & - & 56.3 & -\\
    Flamingo-9B~\cite{NEURIPS2022flamingo} & - & 51.8 & -\\
    BLIP2~\cite{li2023blip2} & - & 65.0 & -\\
    Otter~\cite{li2023otter} & \underline{0.13} & - & -\\
    Shikra-13B~\cite{chen2023shikra} & 0.05 & \textbf{77.3} & \textbf{77.5}\\
    \textbf{Ours-7B-[mix]} & \textbf{0.16} & \underline{75.1} & \underline{75.3}\\

    \bottomrule[1.2pt]
  \end{tabular}}
  \caption{\textbf{Quantitative evaluation} was conducted on both ImageNet-100 and VQAv2 datasets employing a zero-shot approach. The outcomes substantiate that our training strategy exhibits no detrimental impact on the zero-shot performance.}
  \label{tab:zero-shot-harm}
\end{table}
\subsection{Conclusion}
In conclusion, this paper introduces a groundbreaking paradigm of causal-relevant few-shot learning, significantly expanding the capabilities of Multimodal Large Language Models (MLLMs) within the context of single conversations. Through meticulous experimentation and a carefully devised training strategy, we demonstrate that MLLMs can adeptly establish a mapping between ground-truth input-label pairs, thereby acquiring the proficiency to seamlessly generalize this capacity to previously unencountered images and novel concepts. This pivotal advancement propels MLLMs into uncharted territories, enabling them to not only acquire but also apply knowledge in a manner more akin to human cognition.

{\small
\bibliographystyle{unsrt}
\bibliography{aaai24}
}






\end{document}